# Convolutional Neural Shading for High-Quality 3D Reconstruction from Multi-View Images

Juheon Hwang[1], Taewan Kim[2], Heeseok Oh[3], Jiwoo Kang[4*]

[1]Department of Electrical and Electronic Engineering, Yonsei University, Seoul, 03722, South Korea.
[2]Data Science Major, Dongduk Women's University, Seoul, 02748, South Korea.
[3]Department of Applied AI, Hansung University, Seoul, 02876, South Korea.
[4*]Division of Artificial Intelligence Engineering, Sookmyung Women's University, Seoul, 04310, South Korea.

*Corresponding author(s). E-mail(s): jwkang@sookmyung.com;
Contributing authors: consecrated.h@yonsei.ac.kr;
kimtwan21@dongduk.ac.kr; ohhs@hansung.ac.kr;

**Abstract**

We propose a convolutional neural shading (CNS), a novel pipeline to reconstruct high-quality 3D shapes from multi-view images. Several recent studies have used neural radiance fields and other neural differentiable rendering methods to understand 3D geometry. However, these approaches rely on single-point geometric information, such as positions and normals of the surface, leading to a lack of detailed local geometry. Our approach addresses the inherent limitations of single-point information by leveraging a neural shader to capture variations even in dark and textureless regions with a convolutional neural shader, resulting in far more accurate geometry predictions. Additionally, our method mitigates surface irregularities at image boundaries by introducing a fine-detail displacement network, which utilizes spatial information of surface geometry and learns fine displacement details by correlating neighboring values in the rendering coordinates. Through extensive experiments, our proposed method has demonstrated significant quality improvements in the reconstructed shapes and rendered images over current state-of-the-art methods.

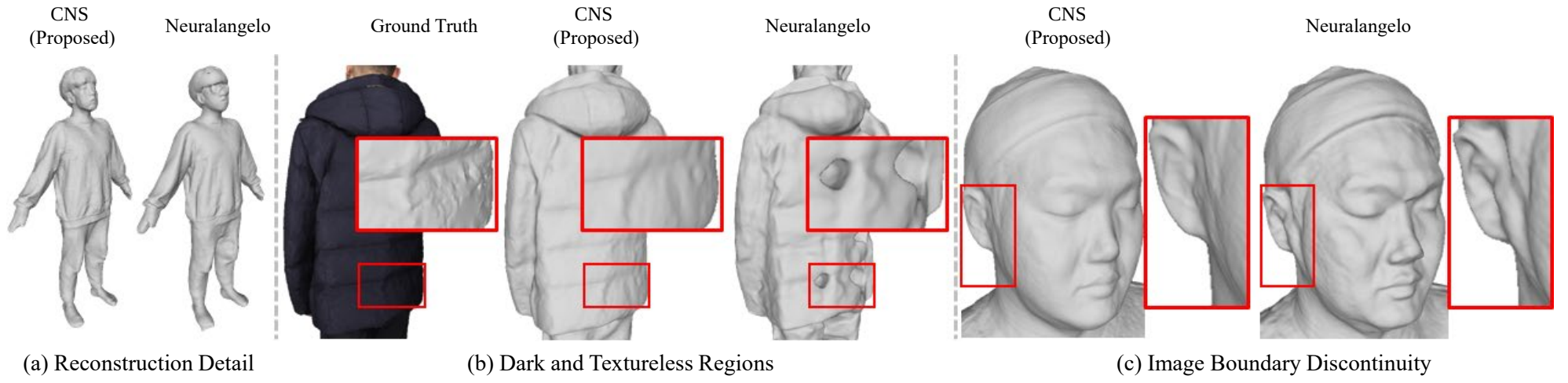

**Fig. 1**: The proposed pipeline for 3D object reconstruction produced high-fidelity surfaces by correlating the spatial information. The comparisons are performed with current state-of-the-art method, Neuralangelo [20]. The proposed method achieved significant improvements over the previous method. Best viewed with zoom-in.

# 1 Introduction

3D reconstruction from multi-view images is an essential topic in the film and gaming industries [1–3]. Traditionally, creating 3D content requires laborious manual work by 3D artists. Numerous efforts [4–7] have been made to reduce the cost of human labor with 3D reconstruction techniques using multi-view images. In past years, 3D reconstruction has been performed using stereo matching techniques [8–10], which compare features between two images to calculate the disparity. However, stereo-matching reconstruction produces several geometrical artifacts due to the accumulated error from the multi-view depth fusion, such as the misaligned surfaces of different viewpoints.

To address this issue, there have been several studies on 3D geometry learning with neural differentiable rendering [11–13]. These studies used variants of Neural Radiance Field (NeRF) [14] technique that models the radiance field of a scene to synthesize novel views. NeRF predicts image colors by sampling points along the ray from the image coordinates to the view direction and learning the volume density and color at each point in 3D space via the volume rendering technique. Meanwhile, NeRF-based 3D reconstruction networks [15–18] use the implicit neural function to represent accurate 3D geometry with a signed distance function (SDF). In [15], 3D geometry is reconstructed without sampling all the points along the ray. Instead, it uses an intersection point of the ray with the surface represented by the implicit neural function. Recent work in [16] has shown impressive performance in handling occlusion using the implicit neural function and modified weights of the NeRF rendering equation. Although NeRF-based approaches can reconstruct high-fidelity surfaces by learning features in 3D volume, these approaches have significant drawbacks, such as high computational costs to train the implicit neural function and difficulty, especially in reconstructing dark and textureless objects.

Current approaches [15, 16, 18–20] for neural differentiable rendering commonly use a multi-layer perception (MLP) for the neural renderer to compute object colors using single-point information such as surface position, surface normal, and view direction. However, predicting the object's color and 3D geometry using single-point information can lead to serious artifacts.

Most of all, it is difficult to learn the correlation between neighboring points with single-point information. This results in a lack of detail in the local geometry of the reconstructed mesh, as shown in Fig. 1 (a). Also, inaccurate mapping between its 3D geometry and color arises in high-frequency detailed geometry, such as wrinkles. Specifically, previous methods show degraded performance in dark and textureless regions, as shown in Fig. 1 (b). Furthermore, the previous methods struggle to clearly estimate the 3D geometry at the image boundary, which causes discontinuity on the surface, as shown in Fig. 1 (c). These issues arise due to the inherent limitation of MLP-based techniques in representing discrete values with continuous input, and thus the results struggle to reflect the neighboring context.

To this end, we propose Convolutional Neural Shading (CNS), a novel pipeline to reconstruct a high-quality 3D object from multi-view images. Our method can represent fine details of the target by correlating neighboring values in geometric and rendered phases, as shown in Fig. 1 (a).

Our method achieves a highly accurate representation of the target by efficiently correlating neighboring values in the 3D geometry and the rendered image, respectively, as shown in Fig. 1 (a). In particular, our method addresses inaccurate geometry problems caused by the lack of information from a single point by proposing a convolutional neural shader that captures local variations and contexts to accurately predict the 3D object's geometry in dark and textureless regions, as shown in Fig. 1 (b). Furthermore, as shown in Fig. 1 (c), our method addresses discontinuities on the surface corresponding to the image boundaries using information on the local geometry by introducing a fine-detail displacement network, which learns displacements of positions and surface normals in the rendering coordinates. Through exhaustive experiments, the proposed method shows significant improvements in reconstructing 3D shapes and rendering images, demonstrating the strong benefits of the proposed method over state-of-the-art methods.

In summary, we propose a 3D reconstruction pipeline that efficiently handles geometry and shading artifacts using the convolutional neural shader, in which

- The accurate 3D geometry and color of the target objects are predicted by correlating spatial information,
- Robust 3D reconstruction at image boundaries is accomplished by leveraging local geometry information, and
- Detail positions and surface normals are represented using fine displacement learning by relating neighboring values in the rendering coordinates.

In the remainder of this paper, we describe the details of our proposed approach and validate its effectiveness through experiments. Section 2 reviews related work in multi-view 3D reconstruction and neural rendering. Section 3 presents our proposed Convolutional Neural Shading (CNS) pipeline, including the fine-detail displacement network and the convolutional neural shader. Section 4 provides experimental results, including comparisons with state-of-the-art methods, ablation studies, and visualizations. Finally, Section 5 concludes the paper with a summary and discussion of future work.

## 2 Related Work

Multi-view reconstruction is a prominent computer vision technique that seeks to reconstruct 3D models or scenes from a 2D image set captured from multiple viewpoints. This approach has been widely researched and developed with applications in areas, such as robotics, virtual reality, film, and gaming industries.

### 2.1 Multi-view Reconstruction

Traditional approaches in multi-view reconstruction generally used the hand-crafted features of the geometry. Structure from Motion [21] (SfM) and Simultaneous Localization and Mapping [22] (SLAM) are two examples of early approaches that can yield 3D reconstructions of acceptable quality. Furthermore, multi-view stereo [23] (MVS) is an extension of stereo matching that uses more than two images to improve the accuracy and completeness of depth estimation. MVS algorithms mostly have a similar structure to stereo matching algorithms, but they can consider consistency among multiple images or use feature-based methods. Plane sweep [24, 25] is another technique that projects a set of images onto a virtual plane at different depths and computes the photo-consistency of each pixel across the images to find the optimal depth. Plane sweep works well for complex 3D scenes and has the advantage of real-time processing.

Recently, deep learning-based methods have been applied to the multi-view reconstruction problem. For example, some methods [9, 26, 27] used convolutional neural networks (CNNs) or transformers [28] to extract and fuse features from multiple views. Unlike these approaches that determine surfaces by feature matching of paired or multiple images, our method renders colors via a neural renderer based on the positions and normals from the estimated surface.

### 2.2 Implicit Neural Reconstruction

One of the pioneering works in this field is Neural Radiance Fields [14] (NeRF), which uses a multi-layer perceptron (MLP) to represent a scene as a continuous volumetric function that maps a 3D location and viewing direction to a volume density and radiance value. However, NeRF does not explicitly model the surface geometry of the scene, and extracting high-quality surfaces from its learned representation is challenging.

To address this limitation, several works have proposed to combine implicit neural representations with surface reconstruction methods. Implicit Differentiable Renderer [15] (IDR) uses an MLP to represent a scene as an implicit surface function that maps a 3D point to an occupancy value and a color value. IDR also learns a differentiable renderer that projects the implicit surface onto the image plane. However, IDR struggles to reconstruct objects with severe self-occlusion or thin structures. Neural Implicit Surfaces by Volume Rendering [16] (NeuS) proposes a novel volume rendering method to train an MLP that represents a scene as a signed distance function (SDF). NeuS can robustly handle complex structures and self-occlusion. Volume Rendering of Neural Implicit Surfaces [17] (VolSDF) extends IDR by introducing an adaptive sampling scheme that focuses more rays on regions near the surface. However, implicit neural representation has the drawback that there are high computational costs to train the implicit network. To address this problem, Neural Deferred Shading [18] (NDS) combines mesh representation instead of implicit neural representation. NDS can significantly reduce computational costs by using the mesh representation, but it is not easy to represent complex objects with fine details. More recently, Neural Unsigned Distance Function (NeuralUDF) [19] uses an unsigned distance function instead of SDF to represent the surface of the object more precisely. Neuralangelo [20] extends a single-resolution grid to multi-resolution 3D hash grids for better detailed representation of the target object.

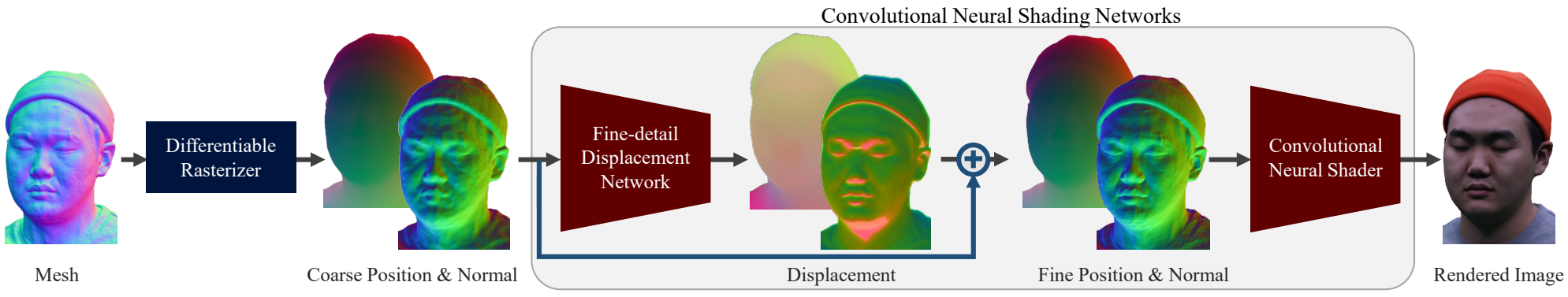

**Fig. 2**: The pipeline of Convolutional Neural Shading (CNS). The rendering procedure is carried out in three consecutive steps: (1) differentiable rasterization, (2) fine-detail displacement learning, and (3) convolutional neural shader. Given multi-view images and camera parameters, the CNS framework jointly optimizes the mesh geometry and neural shading network to reconstruct high-fidelity 3D shapes by capturing fine surface details and spatial color variations. Specifically, the input mesh is projected into image space using differentiable rasterization. We then introduce a displacement network that refines vertex positions to better align with observed image structures. Finally, we propose a convolutional neural shader that predicts pixel-wise colors using local geometry and spatial context, enabling realistic and detailed appearance reconstruction.

Although these methods can extract high-quality surfaces from the given multi-view images, the reconstructed surfaces using them lack the details of local geometry due to the use of single-point information. Therefore, we propose convolutional neural shading to reconstruct the surface more accurately by using the convolutional shading layers to efficiently correlate the neighboring points.

# 3 Convolutional Neural Shading

Fig. 2 describes our proposed convolutional neural shading (CNS) pipeline for multi-view 3D reconstruction. We use the mesh representation instead of the implicit surface representation for the surface of the reconstruction target. The mesh representation allows the network to obtain the positions and surface normals in 2D rendering coordinates via a rasterizer, and thus reduces the computational costs for the implicit neural function training while preserving the details. From multi-view RGB images and the corresponding mask images with camera parameters as input, our pipeline jointly optimizes mesh vertices and the CNS network. Thus, an initial mesh is required for optimization in our pipeline. A 3D primitive can be used as the initial mesh, such as a sphere, cube, or cylinder. Otherwise, if the masks of the captured object are given, we reconstruct a coarse initial mesh from the masks using the shape from the silhouette (SfSil) technique [29]. The CNS network mainly comprises a fine-detail displacement network and a convolutional neural shader for multi-view 3D reconstruction. The fine-detail displacement network learns detail positions and surface normals, respectively, from the barycentric-interpolated geometry in the rendering coordinates. The convolutional neural shader learns the target colors from fine-detail positions and surface normals by correlating spatial information. The mesh is upsampled at regular intervals using the subdivision to progressively increase the resolution of the mesh during the training.

### 3.1 Fine-detail Displacement Network

The precision of the positions and surface normals of the rasterized mesh is highly dependent on the resolution of the mesh. The positions and surface normals of a coarse mesh are projected onto the rendering coordinates using barycentric interpolation. Thus, the positions and surface normals of the coarse mesh cannot accurately represent the geometry in the projected coordinates, leading to inaccurate rendering. The fine-detail displacement network provides high-frequency details of the positions and surface normals, allowing the neural shader to accurately represent the target geometry and color.

The fine-detail displacement network $f_\phi$ can be represented as

$$(\Delta \boldsymbol{p}, \Delta \boldsymbol{n}) = f_\phi\left(\boldsymbol{p}, \boldsymbol{n}\right), \tag{1}$$

where $\Delta \boldsymbol{p}$ and $\Delta \boldsymbol{n}$ are the displacements of the positions and the surface normals, respectively. Thus, fine-detail positions $\boldsymbol{p}_{fine}$ and surface normals $\boldsymbol{n}_{fine}$ are defined with displacements as

$$\boldsymbol{p}_{fine} = \boldsymbol{p} + \Delta \boldsymbol{p}, \boldsymbol{n}_{fine} = \frac{\boldsymbol{n} + \Delta \boldsymbol{n}}{|\boldsymbol{n} + \Delta \boldsymbol{n}|}. \tag{2}$$

### 3.2 Convolutional Neural Shader

The convolutional neural shader $f_\theta$ learns the target color using the fine-detailed 3D positions and surface normals $(\boldsymbol{p}_{fine}, \boldsymbol{n}_{fine})$ obtained from the fine-detail displacement network $f_\phi$. Following [14], we use the positional encoding $\gamma(\cdot)$ to map the position vectors to a high-dimensional space [30].

This encoding method maps low-dimensional inputs (e.g., 3D positions) into a higher-dimensional space using a set of sinusoidal functions at different frequencies. This allows the network to better represent high-frequency details in geometry and appearance, which are critical for accurate shading. We chose this particular method due to its proven effectiveness in capturing fine spatial variations, as demonstrated in NeRF-based rendering tasks [30], and because it enhances the expressiveness of our neural shader in representing subtle surface details.

The neural shader uses the view direction $\boldsymbol{v}$ to consider the specular illumination of the captured real object from the local neighborhood information. Thus, the convolutional neural shader learns to render the color $\boldsymbol{c}$ at the 3D point $\boldsymbol{p}_{fine}$ as

$$\boldsymbol{c} = f_\theta\left(\gamma\left(\boldsymbol{p}_{fine}\right), \boldsymbol{n}_{fine}, \boldsymbol{v}\right). \tag{3}$$

We concatenate the surface normals $\boldsymbol{n}$ and the viewing direction $\boldsymbol{v}$ into the activation of the middle of the layers. Instead of using single-point information on mesh geometry in color prediction, the neural shader learns spatial information, leading to better accuracy in color prediction and surface reconstruction.

## 3.3 Loss Functions

This section describes the loss functions used in our reconstruction pipeline. We train our neural networks while jointly optimizing the mesh. Thus, we use the mesh-based loss function $\mathcal{L}_{mesh}$ and the image-based loss function $\mathcal{L}_{img}$ for our pipeline. Mesh-based losses are defined with respect to the geometry of the mesh. In contrast, image-based losses are defined with respect to the rendered RGB image via our neural shader.

The overall loss functions $\mathcal{L}_{total}$ can be defined as

$$\mathcal{L}_{total}\left(G, \theta; I, M\right) = \mathcal{L}_{mesh}\left(G\right) + \lambda_{img}\mathcal{L}_{img}\left(G, \theta, \phi; I, M\right), \tag{4}$$

where $I$ is the ground-truth image in each view, $M$ is the corresponding mask image, $G$ is the mesh to optimize, $\lambda_{img}$ is the balancing constant between mesh-based and image-based losses, and $\theta$ and $\phi$ are the parameters of the convolutional neural shader and the fine-detail displacement network, respectively.

### 3.3.1 Mesh-based Losses

Mesh-based losses $\mathcal{L}_{mesh}(G)$ consist of Laplacian loss $\mathcal{L}_{lap}(G)$ and normal consistency loss $\mathcal{L}_{normal}(G)$ to regularize the geometry of the mesh as

$$\mathcal{L}_{mesh}\left(G\right) = \mathcal{L}_{lap}\left(G\right) + \lambda_{normal}\mathcal{L}_{normal}\left(G\right), \tag{5}$$

where $\lambda_{normal}$ is the balancing constant.

Laplacian loss, which prevents overfitting at each viewpoint by ensuring the smoothness of the mesh and allows for a stable optimization of the mesh, is defined as

$$\mathcal{L}_{lap}\left(G\right) = \frac{1}{N_v}\sum_{i}^{N_v} || \frac{1}{\mathrm{len}\left(s_i\right)} \sum_{j \in s_i} \left(\boldsymbol{x}_j - \boldsymbol{x}_i\right) ||_2^2, \tag{6}$$

where $s_i$ is the set of neighboring indices of the $i^{th}$ vertex, $\boldsymbol{x}$ is the position of the vertix, $N_v$ is the number of the vertices of the mesh, and $\mathrm{len}(\cdot)$ is the operator to obtain the length of the set.

Normal consistency loss, which measures the cosine similarity between the normals of two faces that share the same edge in a 3D mesh, is defined as

$$\mathcal{L}_{normal}\left(G\right) = \frac{1}{\mathrm{len}\left(s_e\right)} \sum_{i,j \in s_e} \left(1 - \boldsymbol{n}_i \cdot \boldsymbol{n}_j\right)^2, \tag{7}$$

where $s_e$ is the set of pairs of faces that share the edge of the mesh and $\boldsymbol{n}_i$ is the surface normal of the $i^{th}$ face.

### 3.3.2 Image-based Losses

Image-based losses $\mathcal{L}_{img}$ consist of RGB loss $\mathcal{L}_{RGB}$, mask loss $\mathcal{L}_{mask}$, Structural Similarity Index Measure [31] (SSIM) loss $\mathcal{L}_{SSIM}$, and Learned Perceptual Image Patch Similarity [32] (LPIPS) loss $\mathcal{L}_{LPIPS}$ as

$$\begin{aligned}\mathcal{L}_{img}(G,\theta,\phi;I,M) = & \mathcal{L}_{RGB} + \lambda_{mask}\mathcal{L}_{mask} \\ & + \lambda_{SSIM}\mathcal{L}_{SSIM} + \lambda_{LPIPS}\mathcal{L}_{LPIPS},\end{aligned} \tag{8}$$

where $\lambda_{mask}$, $\lambda_{SSIM}$, and $\lambda_{LPIPS}$ are the balancing constants.

RGB loss measures the $\mathbb{L}_1$ distance of RGB values between the ground-truth image $I$ and the rendered image $I'$:

$$\mathcal{L}_{RGB}(I, I') = \|I - I'\|_1 . \tag{9}$$

Our neural shader can learn to render colors from the geometry information of the mesh with RGB loss.

Mask loss ensures that the rasterized area $M'$ of the mesh is consistent with the ground-truth mask area $M$. Thus, we can deform the mesh within the same visual hull as the ground-truth mask. The mask loss is defined with the $\mathbb{L}_2$ distance between the binary masks as

$$\mathcal{L}_{mask}(M, M') = \|M - M'\|_2^2 . \tag{10}$$

LPIPS loss [32] measures the perceptual similarity between intermediate layer's activations of two images for some pre-trained networks. LPIPS, which makes the resulting colors and 3D geometry consistent with human visual perception by catching high-frequency details of 3D geometry, is defined as

$$\mathcal{L}_{LPIPS}(I, I') = \sum_k \tau^k \left(\phi^k(I) - \phi^k(I')\right), \tag{11}$$

where $k$ is the number of activation maps, $\phi^k$ is the $k^{th}$ activations used to measure the LPIPS loss, and $\tau^k$ is its weight.

SSIM loss [31] measures the structural similarity between the rendered image and the ground-truth image. SSIM loss provides a significant advantage in mesh optimization by making the rendered color structurally closer to the ground truth and is defined as

$$\mathcal{L}_{SSIM}(I, I') = 1 - SSIM(I, I') . \tag{12}$$

Although we refer to this as the "SSIM loss", we actually use the dissimilarity form, $1 - SSIM$, as the loss function to minimize structural differences between the rendered and ground-truth images.

# 4 Experiments

## 4.1 Implementation Details

We used Adam optimizer [33] with a learning rate initialized with $1e-4$ to train our network and optimize the vertices of our mesh. The learning rate for the mesh's vertices decreases by half each time the mesh is upsampled using a subdivision method [34]. We use the positional encoding method [30] with four levels to represent mesh positions. The neural shader consists of two separate networks: diffuse and specular shadings. The diffuse shading network predicts the colors merely from the positions. The specular shading network predicts the colors considering observed viewpoints from the view direction, surface normals, and feature vectors produced by the diffuse shading network. We initialized the network weights with the He initialization [35]. The *PyTorch* framework [36] was used to conduct experiments using a single *Nvidia 3090* GPU.

We conducted experiments using a variety of datasets to evaluate the effectiveness and robustness of our method across diverse scenarios, including general 3D objects, high-resolution scenes, and human models. Specifically, we used the DTU dataset [37], BlendedMVS [38], THuman [39], and a custom-captured multiview dataset. The DTU dataset provides multi-view images of 3D objects with 49 or 64 viewpoints and ground-truth point clouds. We followed previous works [17, 18] and used 15 scenes to evaluate geometry accuracy using Chamfer distance. The BlendedMVS dataset offers high-resolution synthetic scenes with up to 2048 × 1536 resolution and dense multi-view images (31–143 views), allowing us to test performance on complex inputs. For human-related 3D reconstruction, we used the THuman dataset. Since it only provides 3D scans, we generated multi-view images using Blender by rendering each mesh from 30 evenly spaced viewpoints at 1024 × 1024 resolution. Additionally, to assess real-world applicability, we constructed a custom dataset using a DSLR-based multi-view camera system, capturing various subjects including faces and clothed human bodies at 1080 × 1920 resolution with 50 views per subject. Together, these datasets provide a comprehensive benchmark for evaluating reconstruction quality under varying lighting conditions, object types, resolutions, and view sparsity.

## 4.2 3D Reconstruction

We evaluate our results qualitatively and quantitatively with state-of-the-art 3D reconstruction methods [15–20]. To demonstrate our method, we compared the results with other recent methods in three different cases. First, for general cases, we reconstructed the object using dense multi-view images. Next, we evaluated our method on dark and textureless objects. Finally, we evaluated the results of challenging cases using sparse-view images.

### 4.2.1 Comparison Methods

We compared the rendered images and the reconstructed meshes from our method with four state-of-the-art methods: IDR [15], VolSDF [17], NeuS [16], and NDS [18], NeuralUDF [19], Neuralangelo [20], and SuGaR [40]. . Also, we compared our method with COLMAP [41] as a baseline for the 3D reconstruction method. These methods use MLPs in the neural renderer to predict colors and 3D surfaces from the geometrical information of the single pixel on the image coordinates. IDR, VolSDF, NeuS, NeuralUDF, Neuralangelo, and SuGaR represent surfaces with implicit neural functions, *i.e.,* either SDF or density.

In particular, SuGaR leverages 3D Gaussian Splatting [42] to extract high-quality triangle meshes efficiently. In contrast, NDS uses a mesh representation similar to our method.

COLMAP [41] is a standard technique for reconstructing 3D points from multi-view images. It uses patch-based stereo matching technology pairwise and integrates them. IDR [15] trains the neural renderer and the implicit neural function by predicting RGB values at points on the surface that intersect with the rays from the image coordinates. VolSDF [17] extends IDR by determining the colors using the volume rendering method in NeRF, which integrates the colors of the sampled points along the ray from the image coordinates. NeuS [16] also uses the volume rendering method in NeRF, but NeuS modifies the NeRF volume rendering equation to use accurate weights depending on the location of the surface when integrating colors along the ray. NDS [18] uses an initial mesh from the masks given and optimizes the mesh while training the neural renderer. NeuralUDF [19] is based on the implicit neural function similar to IDR, but predicts an unsigned distance function (UDF) instead of SDF. Neuralangelo [20] combines neural surface rendering with multi-resolution 3D hash grids for better representation.

SuGaR [40] extracts surface-aligned triangle meshes from 3D Gaussian Splatting by introducing a regularization term and a level-set based Poisson reconstruction.

### 4.2.2 Common Object

For quantitative comparisons, we used the $\mathbb{L}_1$ *Chamfer* distance to validate the accuracy of 3D geometry between the reconstructed meshes and the ground-truth point clouds on 15 scenes from the DTU dataset [37].

Table 1 summarized the results in terms of the *Chamfer* distances (CD) ($\times 10^{-3}$). We also qualitatively compared the rendered images and reconstructed meshes from our method and other baseline methods on the DTU [37] and BlendedMVS [38] datasets with the four best measured methods in Table 1, *i.e.,*, IDR, NeuS, NeuralUDF, and Neuralanglo. For fair comparisons, we made each reconstructed mesh have a similar number of vertices and faces for each scene. Fig. 3 shows qualitative comparisons of rendered images and reconstructed meshes.

The traditional reconstruction method, *i.e.,*, COLMAP, shows the worst performance. The result implies that neural shading-based methods are more effective in reconstructing 3D objects from multi-view images. NDS shows the worst performance in quantitative comparisons among neural shading-based methods, showing the highest *Chamfer* distance of 1.95 due to the loss of geometric information in the rasterization procedure. The IDR, VolSDF, NeuS show comparable quantitative results to each other.. In detail, IDR shows a lower *Chamfer* distance of 0.90 than NDS. Since IDR only uses intersections between the rays from the pixels and the surfaces of the implicit neural function for the rendering, it is prone to deform the surfaces in the implicit function. This results in limited quantitative performance compared to other methods. VolSDF and NeuS show similar measurements, with *Chamfer* distances of 0.86 and 0.84, respectively. Since both methods use point sampling for their rendering, they can estimate the surface geometry more accurately. Specifically, NeuS uses the point sampling method in NeRF volume rendering, where the sampled points do not converge to a single point of the surface. It causes NeuS to fuse inaccurate values from the sampled points. In contrast to NeuS, since the sampled points of VolSDF converge into a single point on the surface using its sampling algorithm, VolSDF can render colors by integrating similar color values. Compared to the methods previously described, it can be seen that NeuralUDF improves the quality of reconstruction using UDF over previous methods with *Chamfer* distances of 0.75. However, it lacks high-frequency details as it learns a single-resolution grid. Neuralanglo learns neural functions in multi-resolution grids, allowing for more detailed representation of 3D shapes with *Chamfer* distances of 0.61. Nevertheless, since it cannot leverage spatially adjacent information, it tends to produce overly rough or smooth surfaces.

**Table 1**: Quantitative comparisons with state-of-the-art methods. The best measure is in **bold**, and the second-best measure is <u>underlined</u>.

| | 24 | 37 | 40 | 55 | 63 | 65 | 69 | 83 | 97 | 105 | 106 | 110 | 114 | 118 | 122 | Mean |
|---|---|---|---|---|---|---|---|---|---|---|---|---|---|---|---|---|
| COLMAP [41] | 0.81 | 2.05 | 0.73 | 1.22 | 1.79 | 1.58 | 1.02 | 3.05 | 1.40 | 2.05 | 1.00 | 1.32 | 0.49 | 0.78 | 1.17 | 1.36 |
| IDR [15] | 1.63 | 1.87 | 0.63 | 0.48 | 1.04 | 0.79 | 0.77 | 1.33 | 1.16 | 0.76 | 0.67 | 0.90 | 0.42 | 0.51 | 0.53 | 0.90 |
| VolSDF [17] | 1.14 | 1.26 | 0.81 | 0.49 | 1.25 | 0.70 | 0.72 | 1.29 | 1.18 | 0.70 | 0.66 | 1.08 | 0.42 | 0.61 | 0.55 | 0.86 |
| NeuS [16] | 1.00 | 1.37 | 0.93 | <u>0.43</u> | 1.10 | 0.65 | 0.57 | 1.48 | 1.09 | 0.83 | 0.52 | 1.20 | 0.35 | 0.49 | 0.54 | 0.84 |
| NDS [18] | 4.24 | 5.25 | 1.30 | 0.53 | 2.47 | 1.22 | 1.35 | 1.59 | 2.77 | 1.15 | 1.02 | 3.18 | 0.62 | 1.65 | 0.91 | 1.95 |
| NeuralUDF [19] | 0.69 | 1.18 | 0.67 | 0.44 | 0.90 | 0.66 | 0.67 | 1.32 | 0.94 | 0.95 | 0.57 | 0.86 | 0.37 | 0.56 | 0.55 | 0.75 |
| Neuralangelo [20] | <u>0.37</u> | <u>0.72</u> | <u>0.35</u> | **0.35** | <u>0.87</u> | <u>0.54</u> | <u>0.53</u> | <u>1.29</u> | <u>0.97</u> | <u>0.73</u> | <u>0.47</u> | <u>0.74</u> | <u>0.32</u> | <u>0.41</u> | **0.43** | <u>0.61</u> |
| SuGaR [40] | 1.47 | 1.33 | 1.13 | 0.61 | 2.25 | 1.71 | 1.15 | 1.63 | 1.62 | 1.07 | 0.79 | 2.45 | 0.98 | 0.88 | 0.79 | 1.33 |
| CNS (Ours) | **0.34** | **0.52** | **0.31** | **0.35** | **0.75** | **0.41** | **0.39** | **1.05** | **0.83** | **0.54** | **0.42** | **0.43** | **0.28** | **0.32** | <u>0.44</u> | **0.49** |

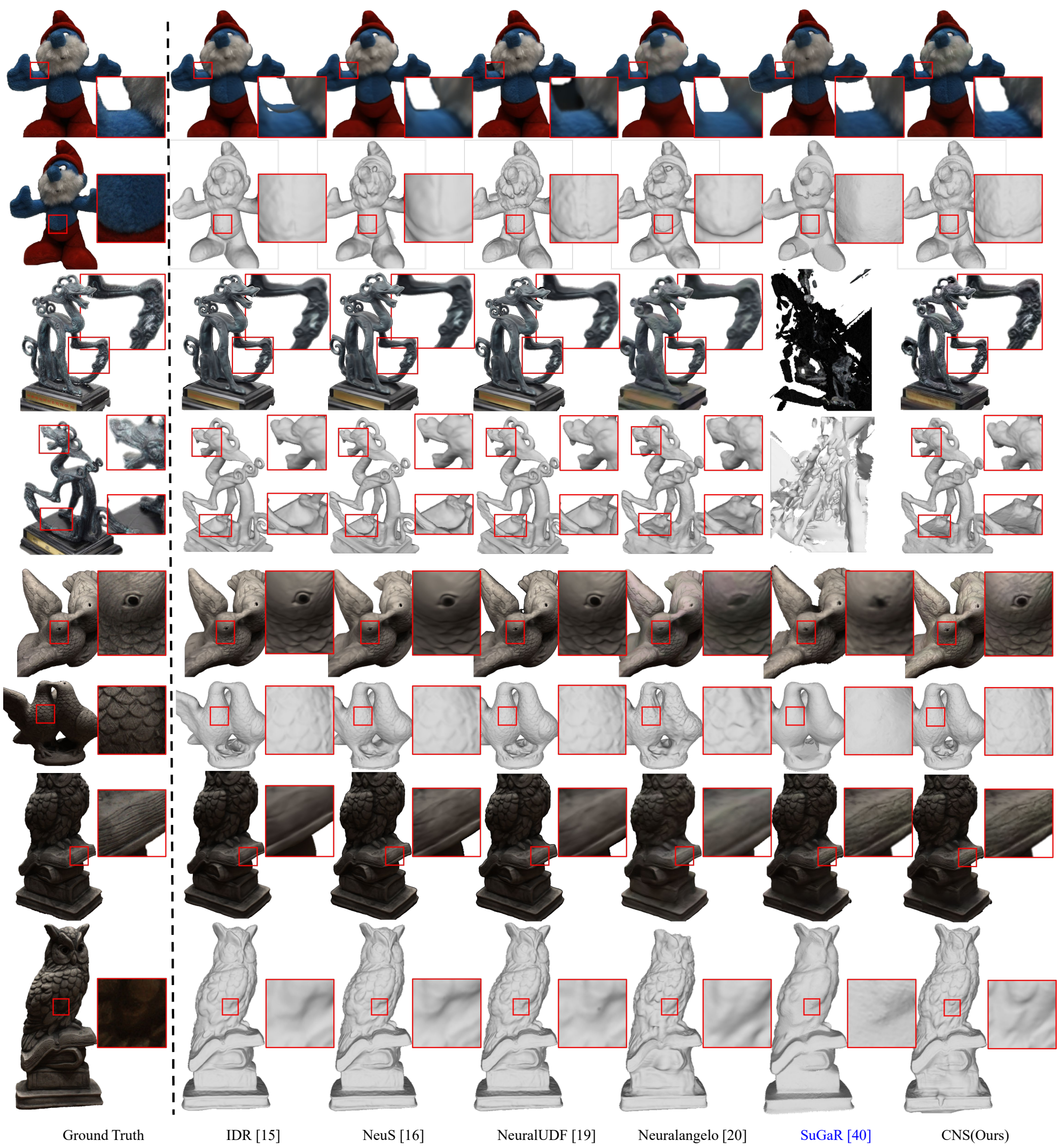

**Fig. 3**: Qualitative comparisons of the rendered images from DTU [37] and Blended-MVS [38] datasets.

**Table 2**: *Chamfer* distance on dark and textureless objects in THuman [39] dataset.

| | *Chamfer Distance* |
|---|---|
| IDR [15] | 1.29 |
| NeuS [16] | 1.01 |
| NeuralUDF [19] | 0.89 |
| Neuralangelo [20] | 0.72 |
| SuGaR [40] | 1.25 |
| CNS (Ours) | **0.41** |

SuGaR [40], a recent method based on Gaussian Splatting, shows a relatively high *Chamfer* distance of 1.33 despite its efficient rendering pipeline. This performance gap is mainly due to its difficulty in reconstructing precise geometry in low-texture and sparse-view settings. SuGaR can produce smooth surfaces but often misses high-frequency structures due to its reliance on sparse Gaussian distributions and limited surface regularization. Compared to the methods used for comparisons, our method achieved promising performance with the lowest *Chamfer* distance of 0.49 with a more accurate surface estimation than other methods, as our convolutional neural shader leverages spatial information. Meanwhile, the fine-detail displacement network provides details of the positions and surface normals obtained from the rasterized mesh. Thus, our method renders more accurate colors with detailed geometry information via more detailed positions and surface normals.

#### 4.2.3 Dark and Texturelss Object

To compare the reconstruction performance on textureless and dark objects, we synthesized multi-view images using clothed human body scans from the THuman [39] dataset by choosing ten dark and textureless clothed human scans. Here, we rendered multi-view images with 30 evenly spaced view directions using Blender.

Fig. 4 shows qualitative comparisons between the reconstructed meshes from the proposed method and current state-of-the-art methods. Furthermore, Table 2 shows quantitative comparisons of the results in terms of the average *Chamfer* distance.

The current state-of-the-art methods struggled to separate the inherent dark color of the object from effects, such as shadows, when estimating the 3D geometry of dark and textureless objects. This is particularly because NeRF-like methods, such as IDR, NeuS, NeuralUDF, and Neuralangelo, tend to inaccurately estimate the geometry of the object's dark areas based on a single ray from each pixel, as shown in Fig. 4.

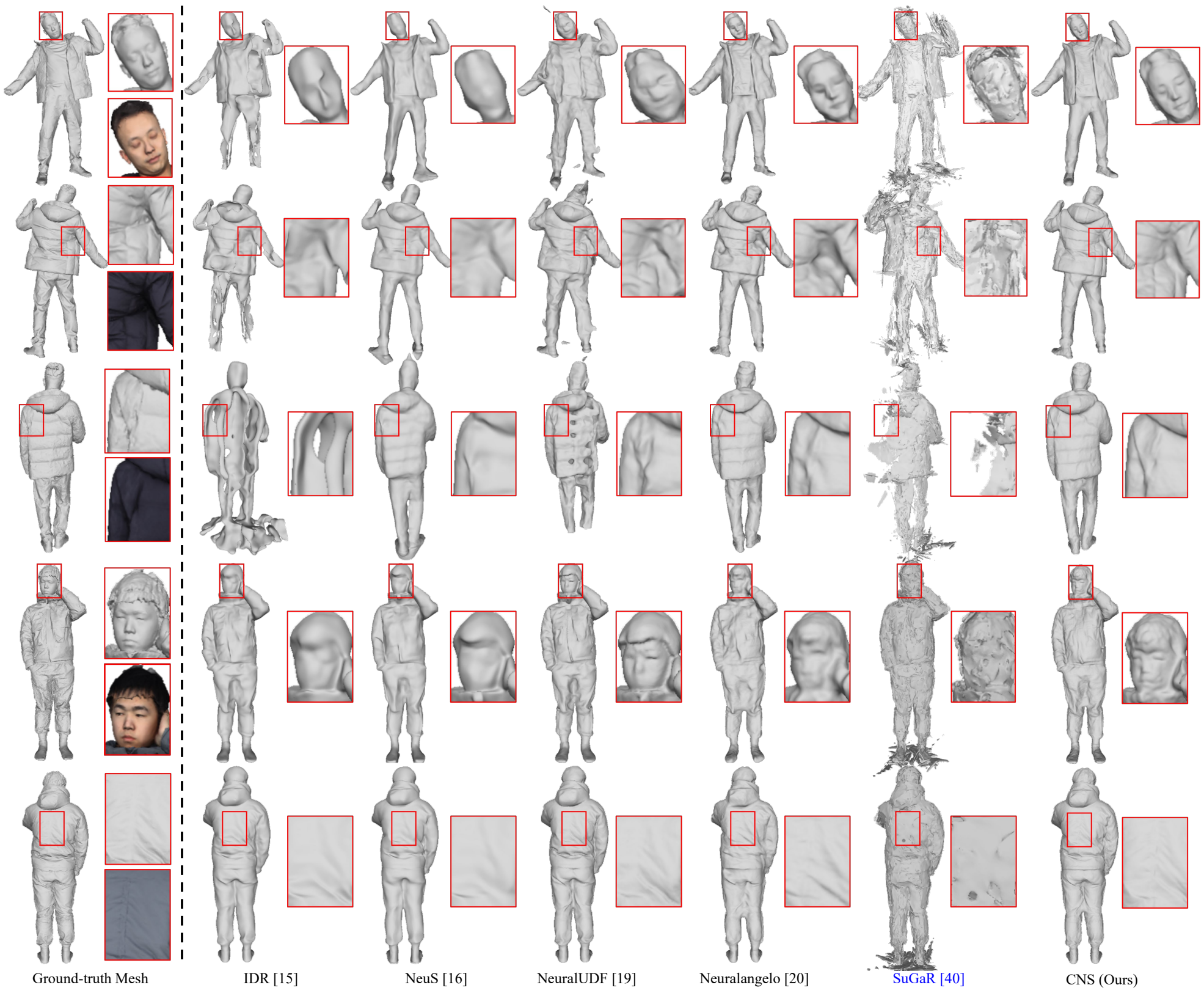

**Fig. 4**: Qualitative comparison on images containing dark and textureless objects.

SuGaR [40], despite its efficient Gaussian-based representation, also suffers from limited accuracy in these challenging regions, showing a high *Chamfer* distance of 1.25. Its reliance on sparse Gaussian primitives and weak surface coupling leads to over-smoothed surfaces that miss important local structures such as sharp wrinkles or silhouette transitions, which are crucial in dark-clothed regions. In contrast, our method accurately represents the subject's face compared to other methods, yielding a mesh closest to the ground-truth mesh. In particular, our method is shown to accurately reconstruct the geometry of dark and textureless regions, such as dark wrinkles. These results are consistent with the *Chamfer* distances in Table 2, where IDR showed the highest distance of 1.29 and our method achieved the lowest distance of 0.41.

### 4.2.4 Sparse view Object

We used sparse viewpoints from eight directions to evaluate the quality of multi-view reconstruction in challenging cases. Fig. 5 shows qualitative comparisons of the reconstructed meshes from our method and current state-of-the-art methods on BlendedMVS [38], THuman [39], and our custom-captured datasets.

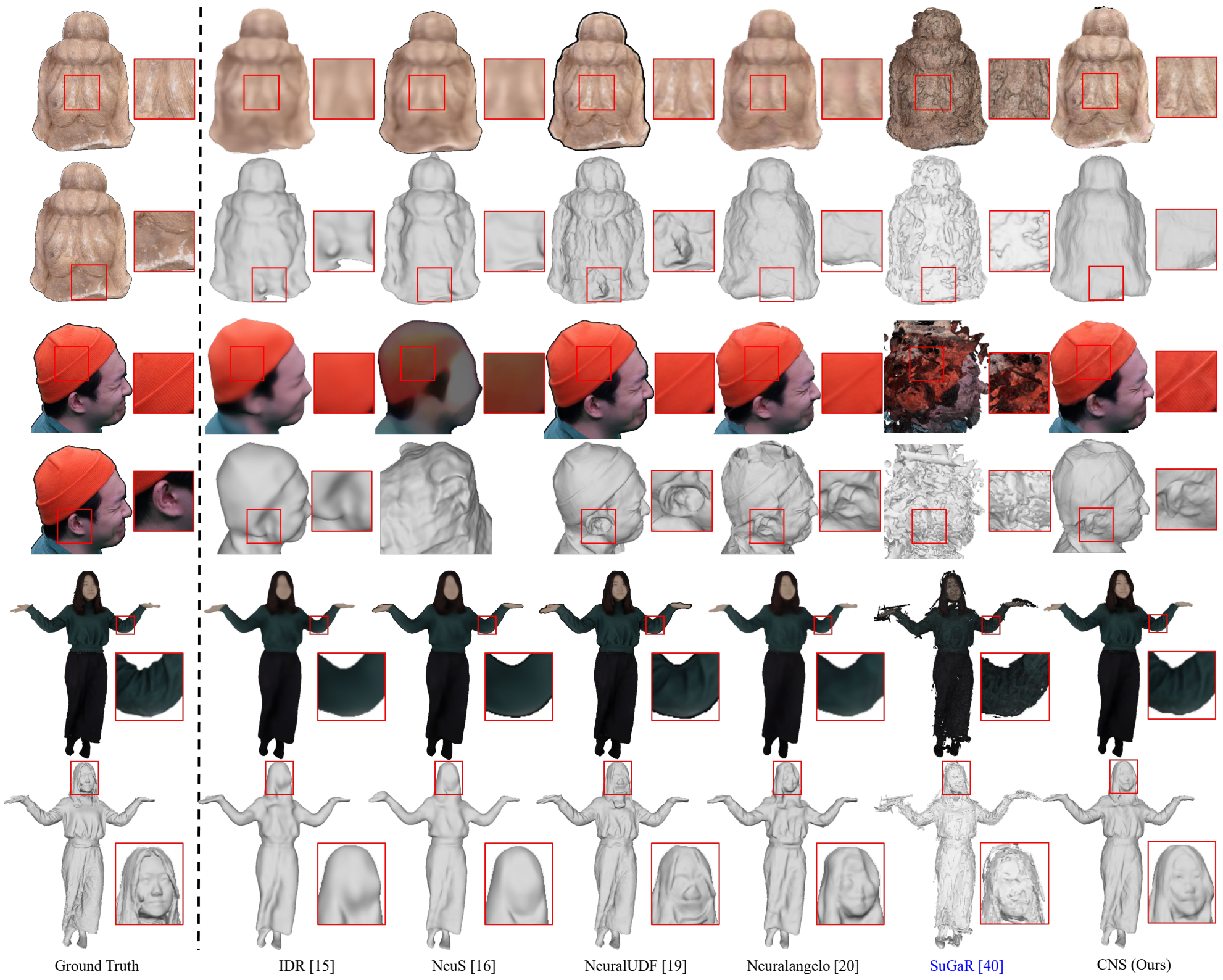

**Fig. 5**: Qualitative result of the meshes in the sparse view reconstruction case (8 views).

In sparse view cases, it is difficult for other NeRF-like methods to accurately learn the surfaces and colors because they train their neural networks to produce the colors and SDF (or density) values from all the points in the rendering volume. In contrast to NeRF-like methods, including Gaussian splatting-based approaches such as SuGaR [40] , the proposed method iteratively updates the initial mesh to most closely represent the given images, which allows for better reconstruction of the target object even in the case of sparse views, allowing for highly accurate surface estimation of the reconstruction target. In particular, our method uses a convolutional neural shader composed of convolutional shading layers, which estimates color based on spatial positions and surface normals. This enables detailed representations even at image boundaries captured from sparse viewpoints.

### 4.3 Error Map Visualization

To further verify the performance of our method, we visualize the Chamfer distance error map of the reconstructed meshes from DTU [37] and THuman [39] datasets in

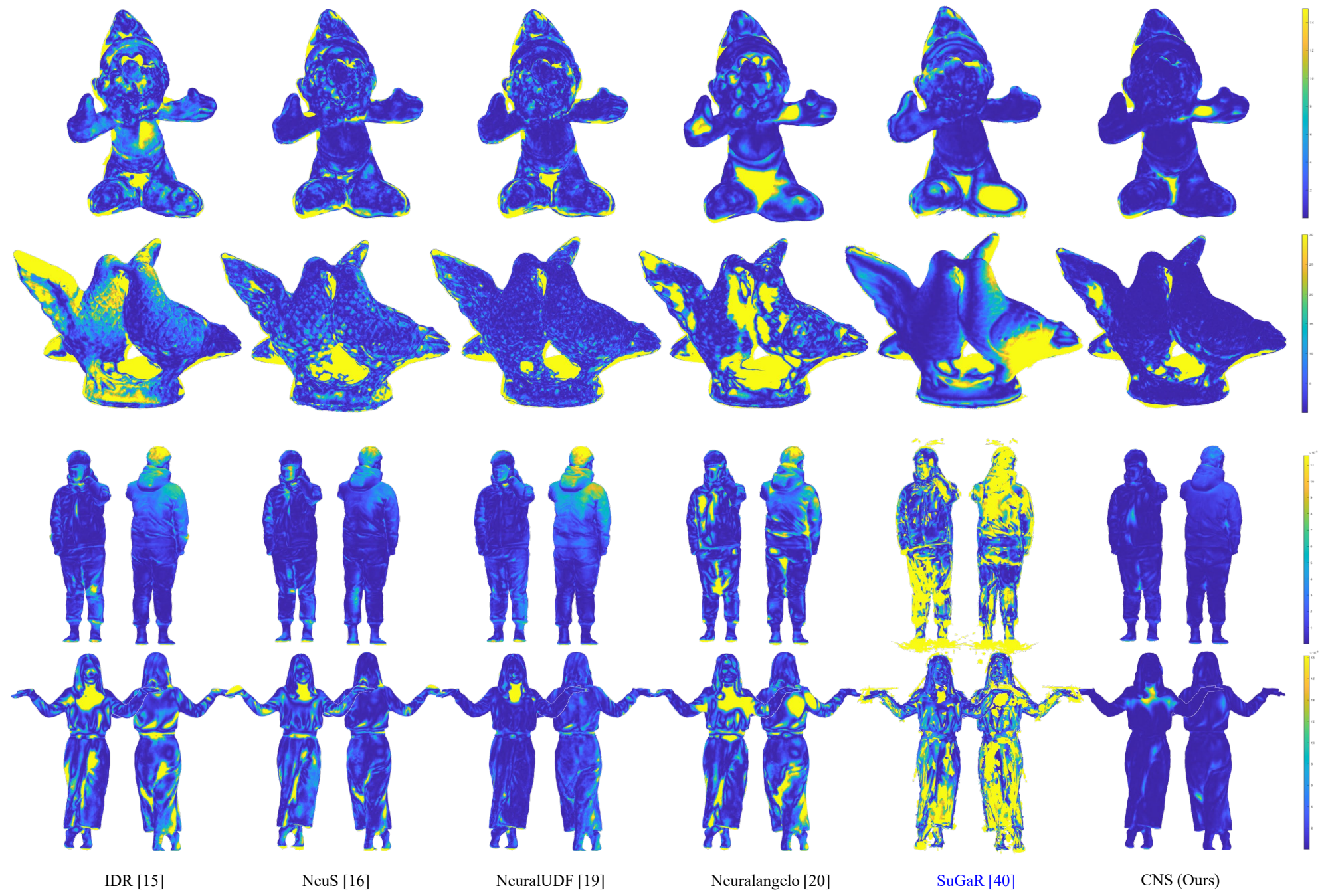

**Fig. 6**: Chamfer distance error map on DTU [37] and THuman [39] datasets.

Fig. 6. We made reconstructed meshes have a similar number of vertices and faces for each scene for fair comparisons. The yellow regions of the surfaces on the error maps represent higher $Chamfer$ distances from the ground-truth meshes, while the dark blue regions represent lower *Chamfer* distances. The results show that the proposed method can reconstruct the surface of the object with low and even errors, demonstrating the fine-detail representation power of our method.

The results of the error map confirm that the performance improvements of the proposed method presented in Tables 1 and 2 come from the representation of its power to represent fine-detail shapes and textures, demonstrating strong benefits of our method in reconstructing 3D objects from multi-view images.

## 4.4 Ablation Study

To validate each of the convolution shading and displacement refinement modules, we qualitatively evaluate the performance of the mesh reconstruction on BlendedMVS [38] and THuman [39] datasets for the ablation study. Fig. 7 shows qualitative comparisons of the rendered images and reconstructed meshes from the ablated configurations and our method. In detail, we replaced entire convolutional shading layers with multi-layer perceptrons and eliminated the fine-detail displacement network, respectively.

Without using the convolutional shading layers, the surfaces and corresponding rendered images are shown to estimate surfaces inaccurately for dark and textureless objects. This result is consistent with the CNS results in Fig. 4. Additionally, the mesh vertices corresponding to the boundaries of the given images generate some sharp wrinkles due to insufficient geometrical information. With the fine-detail displacement network, we can obtain surfaces with fewer artifacts, as it correlates spatial geometry information by using convolutional shading layers. However, without the displacement network, the reconstructed mesh cannot represent surfaces with sufficient details compared to that with the displacement network. These results demonstrate that our convolutional neural shader and fine-detail displacement network complement each other and contribute significantly to reconstructing the mesh from multiview images.

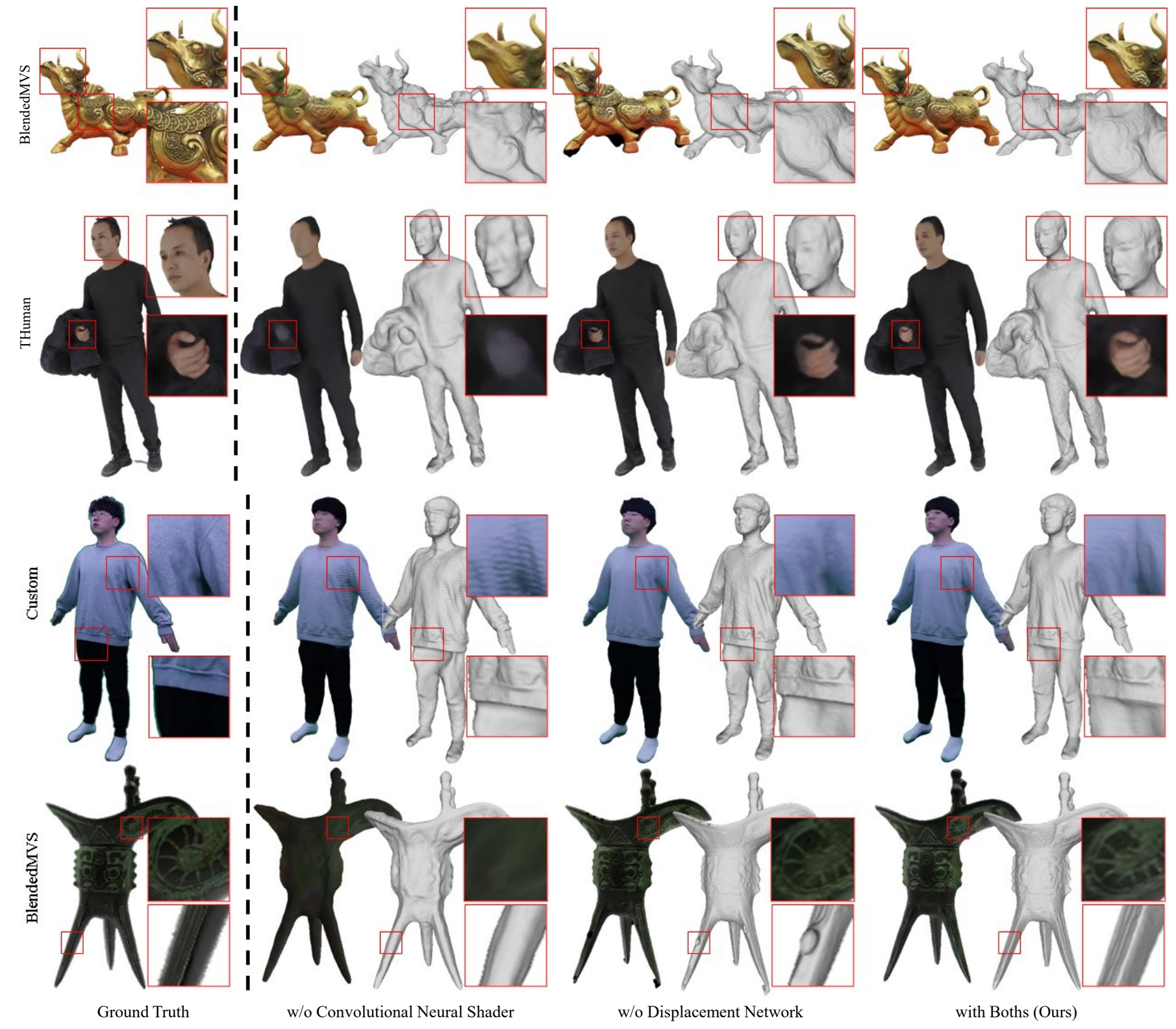

**Fig. 7**: Qualitative comparison of mesh and rendering results of the ablation study.

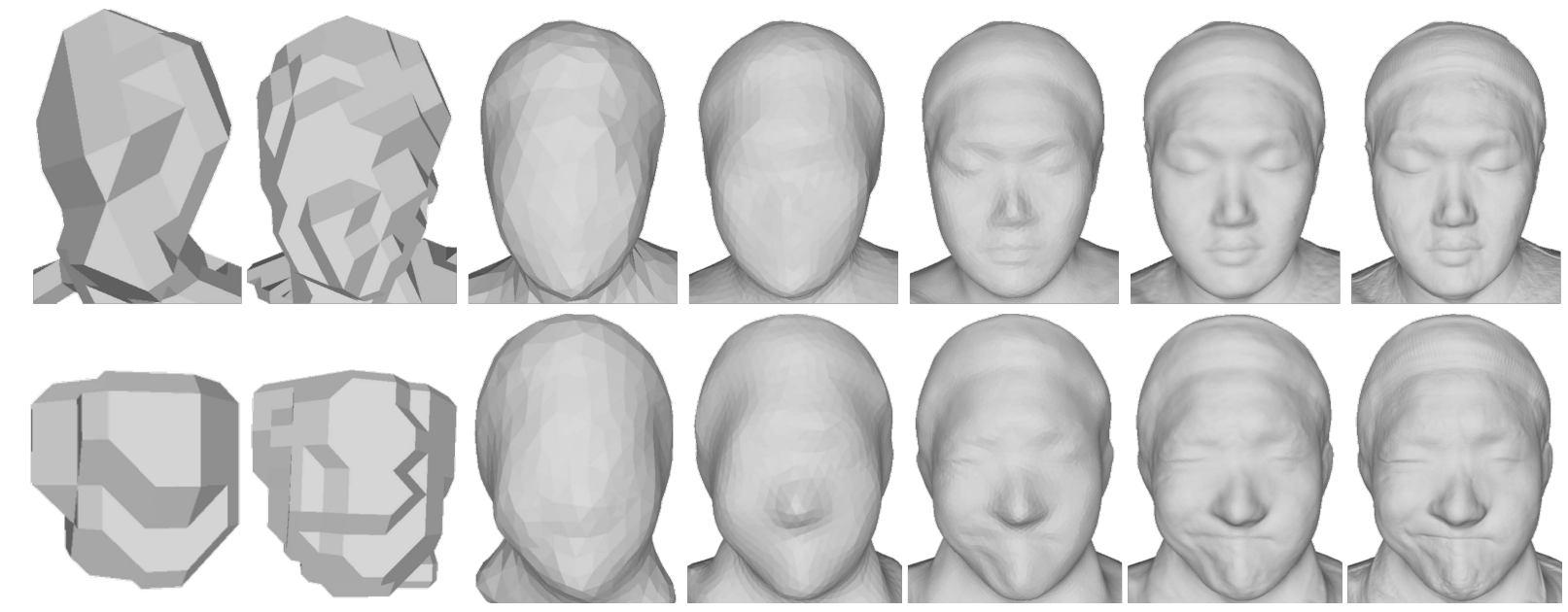

**Fig. 8**: Intermediate meshes in our multi-view 3D reconstruction pipeline.

**Table 3**: Quantitative comparisons of the ablation study.

| | w/o Conv. Shading | w/o Displacement | Ours |
|---|---|---|---|
| *Chamfer* | 1.42 | 0.79 | **0.50** |
| PSNR | 22.68 | 29.19 | **34.29** |

**Table 4**: Comparison between mesh-based and implicit-based surface inputs.

| Method | PSNR ↑ | SSIM ↑ | LPIPS ↓ |
|---|---|---|---|
| Mesh input | **27.35** | **0.912** | **0.087** |
| Implicit input (NeuS) | 25.82 | 0.893 | 0.104 |

For quantitative evaluation, we measure the average *Chamfer* distance and PSNR score for 15 scenes on DTU [37] and BlendedMVS [38] datasets in Table 3. Without using convolutional shading layers, we obtain the *Chamfer* distance of 1.42 and the PSNR value of 22.68 since single-pixel geometry information causes inaccurate color. Without using the fine-detail displacement network, we obtain the *Chamfer* distance of 0.79 and the PSNR value of 29.19 on average. Since the positions and surface normals lack details, the predicted color is less accurate than that predicted by our method. Moreover, a network without a fine-detail displacement network increases the *Chamfer* distance, showing the benefits of the displacement network for reconstructing fine-detail surfaces. By comparing two ablation results, both the convolutional shading layer and the fine-detail displacement network play a significant role in reconstructing object surfaces and rendering target images through the neural renderer.

Additionally, Fig. 8 shows the meshes of intermediate iterations in our pipeline, including the first initial mesh reconstructed with visible masks using the SfSil technique. It can be seen that our method accurately reconstructs meshes in fine detail from even the largely coarse meshes.

To evaluate the benefits of using mesh as input, we conducted an ablation study comparing our approach with a modified version that uses implicit neural functions for surface representation. Specifically, we replaced the mesh input with an implicit field parameterized by a signed distance function (SDF), following the design of NeuS [16]. In this variant, surface normals were computed via differentiating the SDF, and the rest of the pipeline, especially the supervision from multi-view consistency and the refinement network, was kept identical to our original design. Both models were trained under the same conditions on a subset of the DTU dataset [37].

As shown in Table 4, our method using mesh input achieves consistently better reconstruction accuracy. We attribute this improvement to two main factors: (1) mesh surfaces provide explicit and high-quality normal information, which directly benefits the multi-view geometry refinement; and (2) our pipeline leverages mesh connectivity to propagate geometric cues more effectively, whereas implicit fields require significant training iterations and may struggle with local detail preservation, particularly in low-texture regions. These results validate that mesh input is not only efficient but also advantageous for the downstream supervision in our framework.

# 5 Conclusion

We have introduced convolutional neural shading (CNS), a novel pipeline to reconstruct a high-quality 3D object from multi-view images. Our method accurately reconstructs 3D meshes by proposing a convolutional neural shader with multiple convolutional neural shading layers and a fine-detail displacement network to obtain high-quality mesh results. The proposed convolutional neural shader predicted accurate colors from surface positions and surface normals by correlating neighbors, demonstrating our method's effectiveness in challenging cases on textureless objects and sparse-view images. The proposed fine-detail displacement network helped to accurately represent 3D geometry, allowing for fine-detail representation of 3D positions and normals, even in discontinuous regions. The experimental results showed that our method has outperformed current state-of-the-art methods in 3D reconstruction and color estimation.

While our method demonstrates state-of-the-art performance in multi-view 3D reconstruction, it also has certain limitations that suggest directions for future research. In particular, our approach assumes a bounded reconstruction space and struggles to handle unbounded or large-scale scenes with complex backgrounds and significant occlusions, such as those found in Neuralangelo. We observed that when reconstructing scenes where the target object is heavily occluded from certain views or surrounded by diverse background clutter, the quality of the reconstructed mesh degrades significantly. This limitation arises because our method is designed to optimize geometry and appearance within a fixed spatial extent, and does not explicitly handle occlusions or scene-scale variation. Nevertheless, in controlled bounded scenes without significant occlusion, our approach consistently outperforms existing implicit-function-based methods in terms of geometric accuracy and appearance fidelity, as demonstrated across multiple benchmarks. Extending our framework to handle unbounded environments and occlusions, for instance, by integrating visibility modeling or hierarchical spatial reasoning, would be an important direction for future work.

# Declarations

## Author contribution

Juheon Hwang: Investigation, Software, Writing - original draft, Taewan Kim: Writing – review & editing, Heeseok Oh: Writing – review & editing, Jiwoo Kang: Conceptualization, Methodology, Supervision, Writing – review & editing.

## Funding

This work was supported by Institute of Information & communications Technology Planning & Evaluation (IITP) grant funded by the Korea government (MSIT) (No. RS-2023-00229451, Interoperable Digital Human (Avatar) Interlocking Technology Between Heterogeneous Platforms).

## Data availability

We used DTU [37], BlendedMVS [38], and THuman [39, 43] datasets for our experiments. These dataset are publicly accessible online: DTU in https://roboimagedata.compute.dtu.dk, BlendedMVS in https://github.com/YoYo000/BlendedMVS, and THuman in https://github.com/ZhengZerong/THUman4.0-Dataset.

## Conflict of interest

The authors declare no competing interests.

## Ethics approval

Not applicable.